\documentclass[sigconf]{acmart}

\copyrightyear{2021}
\acmYear{2021}
\setcopyright{acmlicensed}\acmConference[MM '21]{Proceedings of the 29th
ACM International Conference on Multimedia}{October 20--24, 2021}{Virtual
Event, China}
\acmBooktitle{Proceedings of the 29th ACM International Conference on
Multimedia (MM '21), October 20--24, 2021, Virtual Event, China}
\acmPrice{15.00}
\acmDOI{10.1145/3474085.3475707}
\acmISBN{978-1-4503-8651-7/21/10}

\usepackage{multirow}

\AtBeginDocument{%
  \providecommand\BibTeX{{%
    \normalfont B\kern-0.5em{\scshape i\kern-0.25em b}\kern-0.8em\TeX}}}

\begin{document}
\fancyhead{}

\title{Decoupled IoU Regression for Object Detection}


%----------------------------------Authors-----------------------------------

\author{Yan Gao}
\authornote{Both authors contributed equally to this research.}
\email{gy243263@alibaba-inc.com}
\affiliation{%
  \institution{Alibaba Group}
  \country{}
}

\author{Qimeng Wang}
\authornotemark[1]
\email{qimengwang@hust.edu.cn}
\affiliation{%
  \institution{Huazhong University of Science and Technology}
  \institution{Alibaba Group}
  \country{}
}

\author{Xu Tang}
\email{buhui.tx@alibaba-inc.com}
\affiliation{%
  \institution{Alibaba Group}
  \country{}
}

\author{Haochen Wang}
\email{zhinong.whc@alibaba-inc.com}
\affiliation{%
  \institution{Alibaba Group}
  \country{}
}

\author{Fei Ding}
\email{feifei.df@alibaba-inc.com}
\affiliation{%
  \institution{Alibaba Group}
  \country{}
}

\author{Jing Li}
\email{lj225205@alibaba-inc.com}
\affiliation{%
  \institution{Alibaba Group}
  \country{}
}

\author{Yao Hu}
\authornote{Corresponding Author.}
\email{yaoohu@alibaba-inc.com}
\affiliation{%
  \institution{Alibaba Group}
  \country{}
}

\begin{abstract}
%\textcolor{red}{your txet}
Non-maximum suppression (NMS) is widely used in object detection pipelines for removing duplicated bounding boxes. The inconsistency between the confidence for NMS and the real localization confidence seriously affects detection performance. Prior works propose to predict Intersection-over-Union (IoU) between bounding boxes and corresponding ground-truths to improve NMS, while accurately predicting IoU is still a challenging problem. We argue that the complex definition of IoU and feature misalignment make it difficult to predict IoU accurately. In this paper, we propose a novel Decoupled IoU Regression (DIR) model to handle these problems. The proposed DIR decouples the traditional localization confidence metric IoU into two new metrics, Purity and Integrity. Purity reflects the proportion of the object area in the detected bounding box, and Integrity refers to the completeness of the detected object area. Separately predicting Purity and Integrity can divide the complex mapping between the bounding box and its IoU into two clearer mappings and model them independently. In addition, a simple but effective feature realignment approach is also introduced to make the IoU regressor work in a hindsight manner, which can make the target mapping more stable. The proposed DIR can be conveniently integrated with existing two-stage detectors and significantly improve their performance. Through a simple implementation of DIR with HTC , we obtain 51.3\% AP on MS COCO benchmark, which outperforms previous methods and achieves state-of-the-art. 
\end{abstract}





\maketitle
\section{Introduction}

   Object detection is a fundamental task in computer vision and it is also the basis for a variety of multimedia applications. The mainstream object detection models can be divided into two-stage methods and one-stage methods according to whether the region proposal stage is included. 
%   Faster R-CNN~\cite{faster} is currently the most popular two-stage method, and many of the recently proposed object detectors~\cite{maskrcnn,cascadercnn,iounet,ltr,double-head} are based on it.  Faster R-CNN first generates region proposals by a Region Proposal Network (RPN)~\cite{faster}, then a classification module distinguishes foreground object proposals from the background proposals and identifies the category of object. The foreground proposals will be further refined by bounding box regression for accurately locating objects. Duplicate detections will be removed by NMS algorithm at last.
   
  In two-stage detectors, the classification scores are used as confidence for NMS to rank candidates. Previous works~\cite{iounet,ltr} indicate that the inconsistency between classification confidence and the actual localization confidence of candidates will cause bounding boxes with higher quality to be removed incorrectly, which will seriously affect the detection performance.

  Predicting IoU or ranking score based on IoU to ameliorate NMS is adopted by many previous works~\cite{iounet,ltr,fcos, goldman2019precise, peng2021systematic, PAA}. However, there is still a large gap between the predicted IoU and actual localization confidence. Accurately localization confidence evaluation is still a challenging problem. We argue that there are two leading causes that limit the performance of current localization confidence prediction models.

      \begin{figure}[t!]
      \begin{center}
      \includegraphics[width=1.0\linewidth]{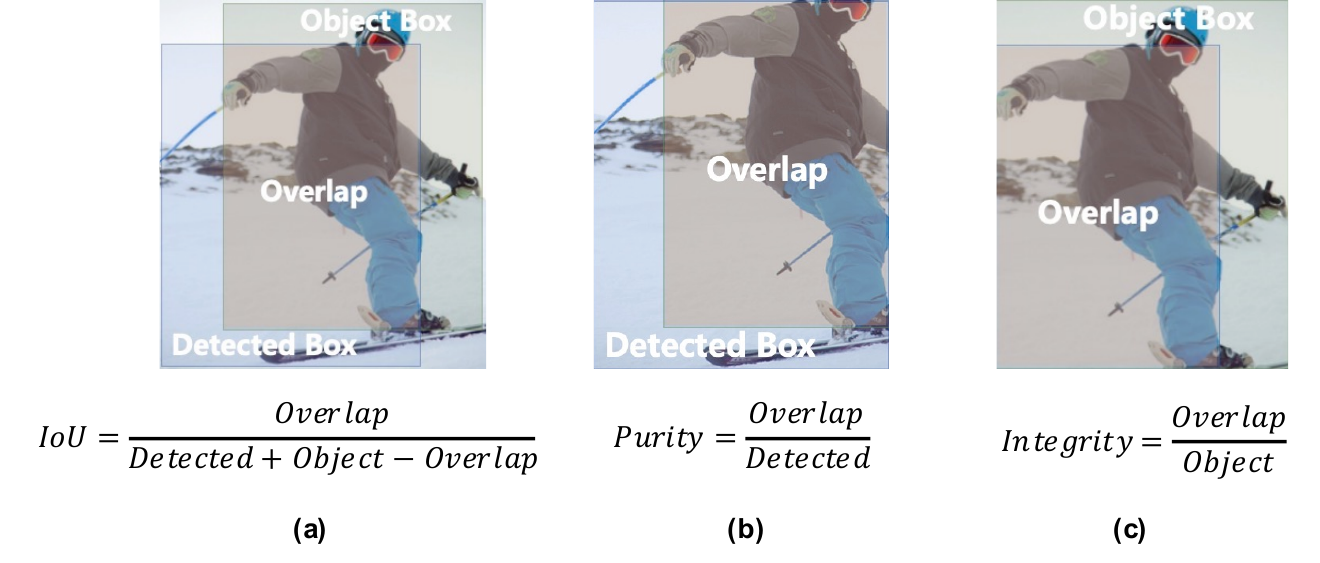}
      \end{center}
          \caption{Definition of IoU, Purity, and Integrity. IoU depends on the area of the intersection and the area of the two boxes. Purity indicates the ratio of the area belongs to the object to the detected bounding box. Integrity indicates the completeness of the object area, which depends on the area of the intersection and the ground-truth object.}
      \label{fig:iou}
      \end{figure}

  The first reason is that the definition of IoU between the bounding boxes and the corresponding ground-truth is complicated. As illustrated in Figure ~\ref{fig:iou}, IoU is defined as the ratio of the intersection between the prediction box and ground-truth to their union, which is determined by the ratio of the object area in the detection bounding box (Figure ~\ref{fig:iou} b), and the completeness of the detected object area (Figure ~\ref{fig:iou} c). In this paper, we define these two values as Purity and Integrity. Predicting IoU requires the network to implicitly perceive purity and integrity simultaneously. However Purity and Integrity focus on different aspects of the detected box, and the information they rely on is also different. As shown in Figure ~\ref{fig:iou}, Purity focuses on the accuracy of the detected bounding box, and it relies on the information of the detected box itself. Integrity focuses on the recall rate of the detection box to the ground-truth object. Besides the detected box itself, perceiving the Integrity also needs context information and sometimes, prior knowledge. Directly regressing IoU which entangled Purity and Integrity in a black box neural network may not be optimal.
  
  We propose to decouple IoU regression as Purity and Integrity. Specifically, we use two sub-network branches to separately model Purity and Integrity and then combine them to get IoU. Compared with directly predicting IoU which entangles Purity and Integrity, separately predicting them can divide the complex mapping between the bounding box and its IoU into two clearer mappings. Each of the mappings will be modeled and supervised independently. Through a simple algebraic transformation as Eq.\ref{eq:pure_inter_to_iou}, the exact IoU can be obtained by Purity and Integrity.

  \begin{figure}[t!]
  \begin{center}
  \includegraphics[width=0.9\linewidth]{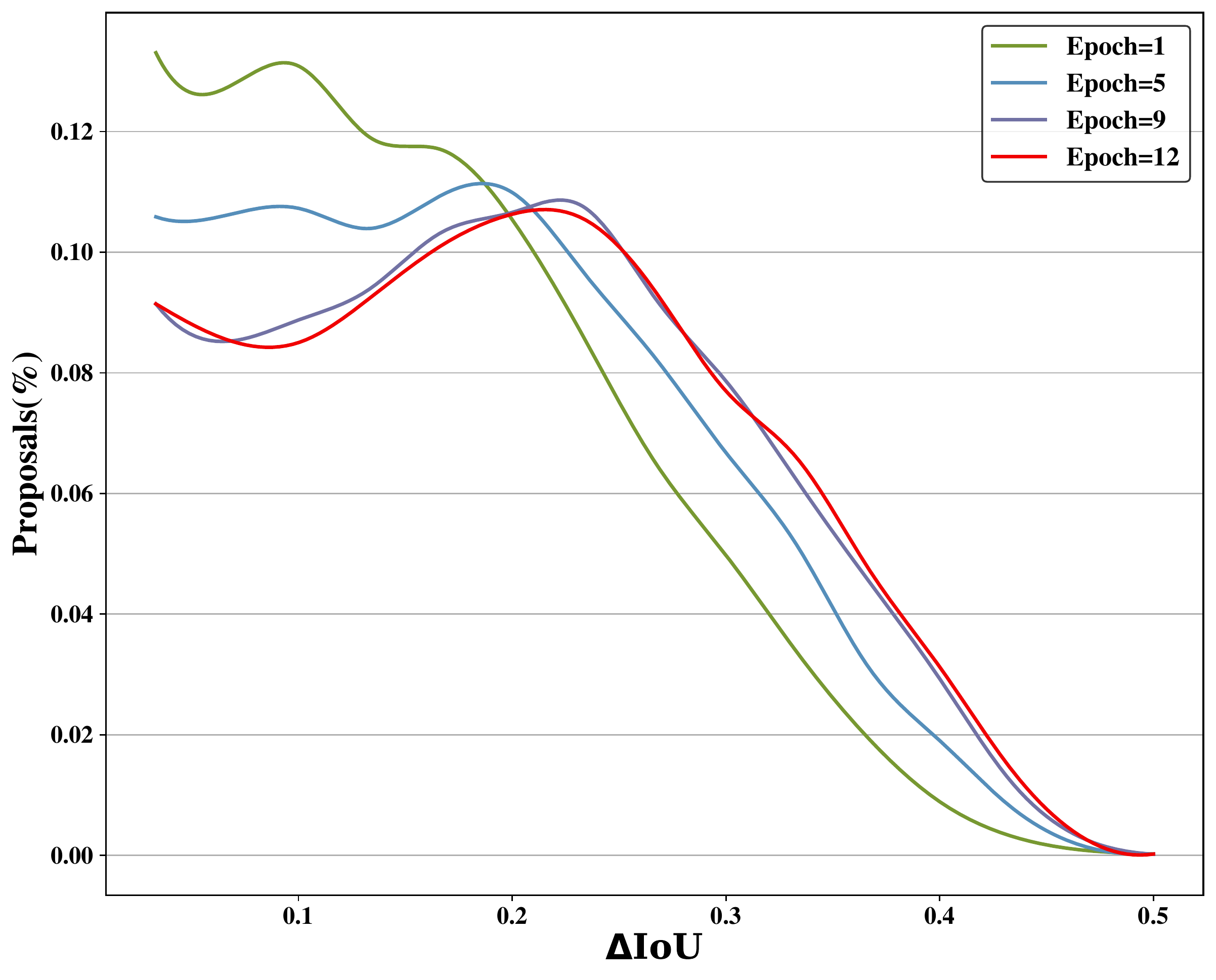}
  \end{center}
      \caption{Distribution of $\Delta IoU$ before and after regression at different epoch. The $\Delta IoU$ axis indicates the IoU difference of proposals and bounding boxes after regression.
      }
  \label{fig:epoch}
  \end{figure} 

   The other issue is the feature misalignment for predicting IoU. Previous works use the feature of a proposal (generated by RPN) to predict the localization confidence of the bounding box which is regressed from the proposal. We study the distribution of IoU difference between the proposals and regressed bounding-boxes by feeding same proposals to detection models at different training stages.
   
   As can be seen in Figure \ref{fig:epoch}, the IoU with ground-truth of the proposal and the regressed bounding box is usually significantly different. Using the feature of proposals to predict the IoU of regressed bounding boxes naturally leads to the issue of feature misalignment. In addition, Figure \ref{fig:epoch} also illustrated that the distribution of IoU difference is constantly changing during the training process, \textit{i.e.} the mapping from proposals to the IoU of regressed bounding boxes is not stable during training process, which indicates that the same proposal may need to be mapped to different IoU values at different training iterations. Such instability of supervisions further increases the difficulty of training an IoU prediction model.

   We propose to predict IoU of bounding boxes in a hindsight manner to handle this issue, \textit{i.e.} predicting the IoU of a bounding box after seeing it. Specifically, we adopt an extra RoI-Align~\cite{maskrcnn} operation to extract the features of predicted bounding boxes instead of using the features of proposals to predict IoU. In such a hindsight manner, the IoU of a bounding box is only related to the visual content of the input bounding box and is not affected by the coordinate change caused by the bounding box regression. Therefore, predicting IoU with hindsight can reduce the ambiguity of the target mapping, and make the learning process more stable. Experiments show that this change can significantly improve the accuracy of IoU prediction. 
   
   During training, we choose to use the positive samples of the classification branch to train the IoU regressor, so that the regressor focuses on sorting the positive samples that cannot be distinguished by the classification score. When performing NMS in inference, we consider both classification confidence and IoU and the geometric average of predicted IoUs and classification scores will be used as the confidence.

   By decoupling IoU into Purity and Integrity and predicting IoU with hindsight, we construct a more powerful localization confidence prediction model: Decoupled IoU Regression(DIR). The proposed DIR can predict more accurate localization confidences of bounding boxes thereby improve the detection performance significantly. Experiments on several state-of-the-art two-stage detectors show that DIR can be conveniently integrated with two-stage detectors and improve their performance. Through a simple implementation with HTC~\cite{chen2019hybrid}, the proposed method under ResNeXt-101-FPN-DCN backbone achieves 51.3\% AP in MS COCO benchmark, which achieves state-of-the-art. The project code and models are released at \textit{https://github.com/qimw/DIR-Net}. 
     %$<$\textcolor{blue}{to be replaced by a link}$>$.

   To summarize, our contributions are listed as follows:
       \begin{itemize}
       
        \item  We propose to handle the complicated IoU mapping between the bounding box and ground-truth by decoupling IoU into Purity and Integrity. Experiments show more accurate IoU can be obtained in such decoupled manner.
       
        \item We point out that feature misalignment can seriously affect the accuracy of localization confidence predictions and propose to predict IoU with hindsight to get more accurate localization confidence.

        \item The proposed DIR can more accurately predict the localization confidences of bounding boxes and significantly improves the performance of multiple object detectors.

    \end{itemize}

\section{Related Works}

% Before deep learning is introduced into object detection, proposals classification with hand-crafted features was the most popular object detection pipeline. Sliding window and other region proposal methods like Selective Search~\cite{selective} and Edgebox~\cite{edgebox} are used to generated proposals. Viola and Jones~\cite{viola} propose to detect human face with Haar-like features and cascaded classifiers. 
\noindent\textbf{Two-stage Detectors.}~In recent years, due to the powerful feature expression ability of convolutional Neural Networks (CNNs), deep CNNs is introduced to object detection\cite{faster,cascadercnn,double-head,ji2019small, wang2019ranking,ltr,yuanqiang2020guided, wu2020forest, han2020exploiting, li2020codan,gao2019c}, Ross \textit{et al} propose RCNN~\cite{rcnn}, which leverages CNNs to extract features of proposals generated by Selective Search and then apply classification and bounding box regression on it. In order to reduce the redundant computations during feature extraction, SPP-Net~\cite{sppnet} and Fast-RCNN~\cite{fastrcnn} propose to extract features from shared feature maps through Spatial Pyramid Pooling and RoiPooling layers respectively. Faster R-CNN~\cite{faster} integrates proposal generation, classification, and bounding box regression tasks into an end-to-end network and gets better performance. Faster R-CNN defines a standard Pipeline for two-stage detectors, many subsequent works~\cite{maskrcnn,cascadercnn,ltr,iounet} are based on it. Mask R-CNN~\cite{maskrcnn} combines semantic segmentation and object detection into a unified network, RoI-Align layer is proposed for better feature extraction. Light-Head R-CNN~\cite{light} introduces a lightweight detection head for faster detection. Cascade R-CNN ~\cite{cascadercnn} constructs a sequence of detection heads to improve detection quality by continuously increasing the positive sample threshold. Double-Head R-CNN proposes to handle classification and regression tasks by different heads. 

\noindent\textbf{One Stage Detectors.}~One-stage detection methods do not need to generate proposals in advance but detect objects from dense locations on the feature maps through Deep CNNs\cite{tang2018pyramidbox,li2019pyramidbox++,liu2020hambox,li2019progressively,liu2020bfbox,chen2019dubox}. Pierre Sermanet \textit{et al}~\cite{overfeat} propose the first one-stage detector OverFeat. YOLO~\cite{yolov1} proposes to predict bounding boxes and class probabilities directly from full images in one evaluation. SSD~\cite{ssd} combines predictions from multiple feature maps to detect objects with different sizes. RetinaNet~\cite{retinanet} proposes focal loss to handle the class imbalance problem of one-stage detectors. Recently, many anchor-free methods~\cite{fcos,extremeNet,centernet,fsaf,cornetnet} that do not require a predefined anchor have been proposed. CornetNet~\cite{cornetnet} proposes to detect objects by finding the corner points of objects and then group them. CenterNet~\cite{centernet} first detects object centers and then predicts objects from those center points. FCOS~\cite{fcos} detects objects by predicting the distance to each boundary of objects. 

\noindent\textbf{Localization Confidence Evaluation Methods.}~The inconsistency between the confidences for NMS and the actual localization quality of bounding boxes will lead to good bounding boxes being wrongly suppressed during NMS. Previous works~\cite{fcos,ltr,iounet,PAA,goldman2019precise,peng2021systematic} propose to predict the localization confidence to alleviate this issue. FCOS~\cite{fcos} proposes to predict the centerness of bounding boxes as the localization confidence. IOU-Net~\cite{iounet} uses an IoU predictor to predict the IoU between proposals and the corresponding ground-truths and leverage the predicted IoU to guide NMS when inference. The IoU predictor in IOU-Net is trained by jittered RoIs generated from ground-truths. Tan \textit{et al} propose a Learning To Rank (LTR)~\cite{ltr} method to predict ranking scores of bounding boxes and fuse it with classification confidence as the final confidence for NMS. 

Our method also contains an IoU predictor to guide NMS. Instead of directly predict IoU~\cite{iounet,PAA,goldman2019precise, peng2021systematic}, we propose to decouple IoU as Purity and Integrity to simplify the IoU mapping between the bounding box and corresponding ground-truth. What's more, we argue that using the features of the proposal to predict the localization confidence of bounding boxes affects the accuracy of these methods. We proposed to predict the localization confidence of the bounding box with hindsight, that is, using the features of the bounding boxes instead of the proposals. This small change can greatly improve the accuracy of IoU prediction. 

\noindent\textbf{Applications of Feature Disentangling in Object Detection.}
Sharing feature between classification and bounding box regression tasks reduce the classification ability of Faster R-CNN. Previous works~\cite{dcrv1,dcrv2,TSD,double-head} propose various feature disentangling methods to decouple classification and regression tasks in object detection. DCR~\cite{dcrv1} proposes to add an extra RCNN classifier that not sharing features with Faster R-CNN backbone to disentangle features. Double Head R-CNN~\cite{double-head} decouples the classification and regression tasks by designing different network architecture for each task. Different from these methods, our work focus on disentangling IoU into two new metrics, and predicting each metric by separate networks. 

\section{Methods}

\begin{figure*}[t!]
\begin{center}
\includegraphics[width=0.9\linewidth]{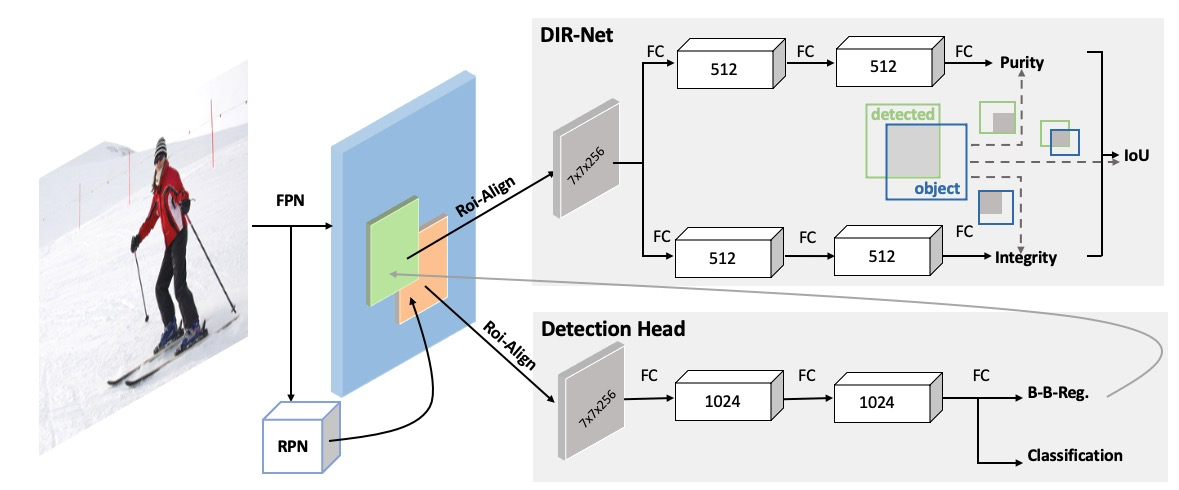}
\end{center}
    \caption{The network architecture of DIR-Net with Faster R-CNN. A CNN backbone with Feature pyramid Network~\cite{fpn} is used to get the feature maps of input image. The standard detection head takes the feature of proposals generated by RPN with a RoI-Align~\cite{maskrcnn} layer and predicts the classification confidence and refines the proposals by bounding box regression. Features of Bounding boxes are extracted by an extra RoI-Align operation and feed to IoU predictor. The DIR-Net predicts the Purity and Integrity and then combines them to obtain the IoU of bounding boxes. }
\label{fig:arch}
\end{figure*}

\subsection{Revisiting Non-Maximum Suppression}
In object detection pipelines, the Non-Maximum Suppression (NMS) algorithm is used to remove duplicate bounding boxes and retain good ones. To achieve this, NMS needs to identify duplicate bounding boxes and rank candidates. Duplicate detections are identified by IoU between bounding boxes. Confidence scores are used to rank candidates. Formally, we denote a set of bounding boxes as \(B=\{b_0,b_1,...b_n\}\) and the confidence scores of bounding boxes as \(C=\{c_0,c_1,...c_n\}\). Staring with the box with the highest confidence, NMS removes all bounding boxes whose IoU with the bounding box is greater than a certain threshold \(\lambda \) and then select the bounding box with the highest confidence in the remaining boxes and repeat the process, until IoU between all boxes is less than \(\lambda \). In the process of NMS, if a inferior bounding box has higher confidence, not only the poor bounding box will be retained, but also the good bounding boxes beside it will be removed by mistake. Therefore, whether the confidence used for ranking candidates in NMS can accurately reflect the quality of the bounding boxes is the key for accurate object detection.

\vspace{-2mm}
\subsection{Decouple IoU as Purity and Integrity}
IoU is widely used in object detection as a metric to measure the similarity of two bounding boxes. The IoU between the detected box and ground-truth reflects the localization quality of the detected box. Let \(b\) denote the detected bounding box and \(g\) denote the corresponding ground-truth. The area of two bounding boxes is \(a_1\), \(a_2\) respectively. The area of overlapping region of \(b\) and \(g\) is defined as \(overlap\). The IoU between \(b\) and \(g\) is defined as:. 
\begin{equation}
    IoU(b, g) = \frac{overlap}{a_1+a_2-overlap}
\label{eq:iou1}  
\end{equation}

Through algebraic transformation, we can get another expression of IoU as:

\begin{equation}          
    IoU(b, g) = \frac{1}{a_1/overlap + a_2/overlap - 1}
   \label{eq:iou2}        
\end{equation}

From Eq.\ref{eq:iou2} we can see that IoU is determined by \(overlap/a_1\) and \(overlap/b_2\). We define these two values as Purity and Integrity, respectively, as in Eq.\ref{eq:pure} and Eq.\ref{eq:integrity}
\begin{equation}        
  \label{eq:pure}
    Purity(b) =\frac{overlap}{a_1}
\end{equation}
\begin{equation}        
  \label{eq:integrity}
    Integrity(b) = \frac{overlap}{a_2}
\end{equation}             

By combine  Purity and Integrity we can get the exact IoU as:

\begin{equation}
    IoU = \frac{1}{1/Purity+1/Integrity-1}
  \label{eq:pure_inter_to_iou}
\end{equation}

Purity reflects how much area in the bounding box belongs to the ground-truth object, and the Integrity reflects the proportion of detected objects part to the entire object. From Eq.\ref{eq:pure_inter_to_iou} we can see that IoU mathematically entangles Purity and Integrity. In our method, we propose to obtain IoU by predicting the Purity and Integrity of a bounding box instead of directly predicting the IoU. Separately predicting Purity and Integrity can divide the complex mapping between the bounding boxes and its IoU into two clearer mappings and model them independently. What's more, by obtaining IoU through Purity and Integrity, a part of the internal structure of the complex mapping of IoU is explicitly defined in the prediction process of Eq.\ref{eq:pure_inter_to_iou}, which further simplify the learning process.

\begin{figure*}[t!]
  \begin{center}
  \includegraphics[width=1.0\linewidth]{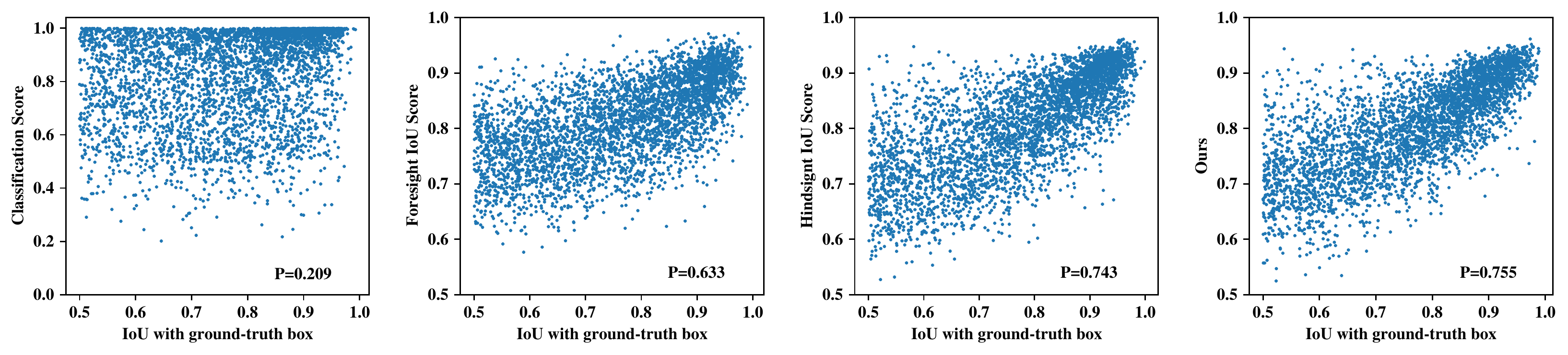}
  \caption{The correlation between the IoU of bounding boxes with the matched ground-truth and the predicted classification/IoU score of different methods. $P$ indicates the Pearson correlation coefficients. }
  \label{fig:compare_p}   
  \end{center} 
\end{figure*} 

\subsection{Hindsight IoU Regression}
 As mentioned before, previous methods propose to predict the localization confidence of a bounding box through features of the corresponding proposal. We denote proposal as \(p\), the bounding boxes after regression as \(b\), and the corresponding ground-truth as \(g\). The actual IoU bwtween \(b\) and \(g\) is defined as \(IoU^*(b,g)\).  The IoU predictor needs to learn the mapping defined as Eq.\ref{eq:noalign} 
, where \(f\) is the feature extractor.
\begin{equation}
    f(p) \rightarrow IoU^*(b,g)
  \label{eq:noalign}
\end{equation}

 Predicting the IoU of bounding box \(b\) through features of the proposal \(f(p)\) requires predicting the localization confidence of \(b\) without seeing it. During the training process, the bounding box \(b\) changes with the training of the bounding box regression, resulting the mapping defined as Eq.\ref{eq:noalign} to change constantly.

In this paper, we define a more stable mapping to model the IoU prediction problem. As defined in Eq.\ref{eq:realign}, we use the features of bounding box \(f(b)\) instead of the features of bounding box \(f(p)\). In this mapping, the IoU of a particular bounding box is stable and only relies on the bounding box itself. Compared with predicting IoU through features of proposals, Our method predicts IoU in a hindsight manner. The target mapping defined in Eq.\ref{eq:realign} is more stable than it in Eq.\ref{eq:noalign} and easier to learn.

 \begin{equation}
    f(b) \rightarrow IoU^*(b,g)
  \label{eq:realign}
 \end{equation}

  We define the Purity and Integrity regressors in our method as \( P(f(b)|\theta_{p}) \) and \( I(f(b)|\theta_{i}) \), where \(\theta_{p}\) and \(\theta_{i}\) is the parameter of regressors. Combining purity and integrity, the proposed Decoupled IoU Regression model in our method is as 

  %\begin{equation}
  % IoU(b_i,g_i) = \frac{P(f(b_i)|\theta_{p}) \cdot I(f(b_i)|\theta_{i})}{P(f(b_i)|\theta_{p})+I(f(b_i)|\theta_{i})-P(f(b_i)|\theta_{p}) \cdot I(f(b_i)|\theta_{i})}
  %\label{eq:hindsight_predictor}
 %\end{equation}

  \begin{equation}
   IoU(b,g) = \frac{1}{1/P(f(b);\theta_{p})+1/I(f(b);\theta_{i})-1}
  \label{eq:hindsight_predictor}
 \end{equation}

\subsection{Implementation with Faster R-CNN}

In this section, we will describe how to implement a simple DIR-Net and integrate it with the popular two-stage detector Faster R-CNN. As shown in Figure \ref{fig:arch}. The DIR-Net contains two separated regression branches for predicting Purity and Integrity respectively. Each of those branches consists of two fully-connected layers with ReLU activation and a fully-connected layer with Sigmoid activation. In the forward process, the features of predicted bounding boxes will be extracted through an extra RoI-Align operation. Then the DIR-Net takes these features to predict the Purity and Integrity for each bounding box and then obtains IoU by Eq.\ref{eq:pure_inter_to_iou}.

During training, all bounding boxes generated by positive proposals will be used to train the DIR-Net. Binary cross entropy loss is used to optimize the regressors. We denote the predicted Purity, Integrity of the bounding box \(b_i\) as \(s_i\), \(t_i\), respectively. The predicted IoU \(c_i\) of bounding box \(b_i\) can be obtained from Eq.\ref{eq:pure_inter_to_iou}. 

The loss functions for Purity, Integrity and IoU are:

\begin{flalign} 
  \begin{split}
    &L_{Puri} = -\frac{1}{N}\sum_{i=1}^{N}Purity^*(b_i)\cdot \log(s_i)+ (1-Purity^*(b_i))\cdot \log(1-s_i) \\
    &L_{Inte} = -\frac{1}{N}\sum_{i=1}^{N}Inte^*(b_i)\cdot \log(t_i)+(1-Inte^*(b_i))\cdot \log(1-t_i) \\
    &L_{IoU} = -\frac{1}{N}\sum_{i=1}^{N}IoU^*(b_i)\cdot \log(c_i)+(1-IoU^*(b_i))\cdot \log(1-c_i)
  \end{split}
\end{flalign}

We use SGD optimizer to train the network end-to-end by combining the loss of Faster R-CNN and DIR as:
\begin{equation}
  L_{Hindsight-Net} = L_{Cls} + L_{Bbox} +L_{Puri} + L_{Inte} +L_{IoU}
  \label{eq:loss_network}
\end{equation}

During inference, the geometric mean of the predicted IoU and classification confidence will be used as the ranking criterion for NMS. To speed up the inference, we only predict the IoU of bounding boxes whose classification confidence is above a certain threshold \(a\).

\section{Experiments}

\begin{table*}[t!]
\begin{center}
    \caption{Comparison between the proposed DIR with existing NMS algorithms on MSCOCO validation set, all models are based on ResNet-50 backbone}
    \begin{tabular}{l c c c c c c c c c}
         \toprule
        Method & Soft-NMS~\cite{softnms} & IoU-NMS~\cite{iounet} & DIR NMS (ours) & AP & \( AP^{50} \) &\(AP^{75}\) & \(AP^{S}\) & \(AP^{M}\) & \(AP^{L} \) \\ \hline
       
        \multirow{4}{*}{Faster R-CNN~\cite{faster}} & &  &        &36.4 & 58.0 & 39.3 & 21.4 & 40.3 &46.7  \\
                  ~    & \checkmark &  &                     &36.9 &\textbf{58.4}& 40.1& 21.9& 40.7& 47.1  \\
          ~& & \checkmark & &37.3 &56.0 &- &- &- &-   \\
          ~& & & \checkmark &\textbf{38.9} & 58.2 & \textbf{42.5} & \textbf{22.6}&\textbf{42.9}&\textbf{51.0} \\ 
        \midrule
        \multirow{4}{*}{Mask R-CNN~\cite{maskrcnn}}  & & &                      &37.3 &59.1 &40.3 &22.0 &40.9 &48.2 \\
        ~ &\checkmark & & & 37.8 & 59.1& 41.3& 22.2 &41.6 & 48.7 \\
        ~ & &\checkmark & & 38.1 & 56.4 & -& - &- & - \\
        ~ & & &\checkmark  & \textbf{39.5} &58.5& \textbf{43.0} &\textbf{23.2} &  \textbf{43.1}&  \textbf{51.9}\\
        \midrule
         \multirow{4}{*}{Cascade R-CNN~\cite{cascadercnn}}  & & & &  40.3 & 58.6& 43.9& 22.9& 43.8& 53.2 \\
        ~ &\checkmark & & & 41.0 & 58.8 & 45.2 & 23.2 & \textbf{44.6} & 54.0\\
        ~ & &\checkmark & & 40.9 & 58.2 & -& - &- & - \\
        ~ & & &\checkmark  & \textbf{41.1} & \textbf{58.8} & \textbf{45.5} & \textbf{23.3} & 44.4  &\textbf{54.8}\\
        \bottomrule 
    \end{tabular} 
    \label{table:compare_with_baselines}
\end{center}
\end{table*}

\subsection{Datasets and Evaluation Metrics}
We conduct experiments on the challenging MS COCO~\cite{mscoco} dataset to evaluate the proposed method. Following common practice~\cite{fcos,faster}, we use the trainval35k split (80k images from train split and a random 35k subset from the val split) for training, and get the ablation study results by evaluating on the minival split (the remaining 5k images from val split). We also report our main result on the test set (20k images) by uploading our detection results to the evaluation server of MS COCO. Average Precision (AP) is exploited as the metric for evaluation.

\subsection{Implementation Details}
Our implementation is based on the MMdetection~\cite{mmdetection} framework.  ResNet-50 and ResNet-101~\cite{resnet} with Feature pyramid network (FPN)~\cite{fpn} are exploited as backbone to extract features of input Images. We train all models over 8 GPUs with synchronized stochastic gradient descent (SGD). The mini-batch size are set to 16 with 2 images per GPU. The initial learning rate is set to 0.02, the momentum and weight decay is set to 0.9 and 0.0001, respectively. Unless specified, the total training iterations is set to 12 epoch. The learning rate is reduced at 8th and 11th epochs. We adopt linear warming up strategy at the first 500 iterations with 0.33 warming up ratio. All the regressors, include Bounding Box Regressor and Decoupled IoU Regressor in our model, are class-agnostic for simplicity. Following Faster R-CNN, we use Cross-Entropy loss for object classification and Smooth L1 loss for bounding box regression. 

\subsection{Data augmentation} 
For convenience, we only use random horizontal flipping for augmentation when training. The shortest side of images is set to 800 and the longest image side is caped by 1333 for both training and testing. We only adopt single scale testing without any augmentation for testing.

\subsection{Ablation Study}
We conduct comprehensive experiments on ResNet-50 backbone and evaluated on COCO minival split to demonstrate the effectiveness of the proposed DIR model and analyze the influences of each module in it. 

\begin{table}[t]
\begin{center}
    \caption{Comparison between the proposed DIR with several strong baselines on MSCOCO validation set, all models are based on ResNet-50 backbone.}
    \begin{tabular}{c c   c  c c  c c  c}
    \toprule
    Method  & \(AP\)& \( AP^{50} \) &\(AP^{75}\) & \(AP^{S}\) & \(AP^{M}\) & \(AP^{L} \) \\
    \midrule 
    DCN~\cite{deformable} &40.0 &62.0 &43.3 &24.0 &43.8 &52.2 \\
    Double-Head~\cite{double-head} & 39.8 & 59.6 & 43.6 &22.7 & 42.9 &53.1 \\
    TSD~\cite{TSD} &40.9 &61.9 &44.4 & 24.2 &44.4 &54.0 \\
    
    \midrule
    DCN + Ours  &41.8 &61.8 &45.6 &25.0 &45.8 &55.6 \\
    Double-Head + Ours  &41.3 &60.6 &54.9 &22.6 & 43.0 &55.1 \\
    TSD + Ours  &42.6 &61.9 &45.6 &25.3 & 46.4 & 55.8 \\
    \bottomrule
    \end{tabular} 
    \label{table:compare_with_strong_baselines}
\end{center}
\end{table}

\begin{table*}[t]
\begin{center}
    \caption{Ablation study experiments results. \textit{Foresight IoU} indicates predict IoU with the features of proposals. \textit{Hindsight IoU} means predict IoU with the feature of bounding boxes. \(P\&I\) indicates obtaining IoU through predicting Purity and Integrity. \(3 \times lr\) indicates train IoU regressor with 3 times learning rate.}
    \begin{tabular}{l c c c c c c c }
         \toprule
         Method &FPS & AP & \( AP^{50} \) &\(AP^{75}\) & \(AP^{S}\) & \(AP^{M}\) & \(AP^{L}\) \\ \hline
         Faster R-CNN & 13.6                       &36.4 & 58.0 & 39.3 & 21.4 &40.3 &46.7  \\
         Faster R-CNN+\textit{Foresight IoU}& 12.1        &37.2 & 57.8& 40.3 &21.5 & 41.1 &48.4 \\
         Faster R-CNN+\textit{Foresight IoU} + P\&I &12.1   &37.9 & 58.1 &41.2 &22.0 & 41.6 & 49.2                     \\
         Faster R-CNN+\textit{Hindsight IoU} & 11.7        &38.3&57.9 & 41.7 & 21.9 &42.1 &50.8                \\
         Faster R-CNN+\textit{Hindsight IoU+ \(3\times lr\)}  &  -      &38.4&57.8 & 42.0 & 22.1 &41.9 &49.9                 \\
         Faster R-CNN+\textit{Hindsight IoU} + P\&I & 11.7 &\textbf{38.9} & \textbf{58.2} & \textbf{42.5} & \textbf{22.6}&\textbf{42.9}&\textbf{51.0}\\ 
    
        \bottomrule 
    \end{tabular} 
    \label{table:ablation}
\end{center}
\end{table*}

\begin{table}[t]
\begin{center}
    \caption{The influence of features used for IoU prediction in training and testing. ROI indicates using the features of proposals and Bounding box indicates using features of Bounding boxes. Training and Inference indicates the stages of using the corresponding features. All experiments are conducted with Faster R-CNN.}
    \begin{tabular}{l c c c c c c c}
         \toprule
         Training & Inference & AP & \( AP^{50} \) &\(AP^{75}\) \\  \hline
          - &  - &          36.4 & 58.0 & 39.3 \\
         ROI  & ROI           &37.9 & 58.1 &41.2\\
         ROI  & Bounding box   &38.3 &58.0 & 41.7\\
         Bounding box  & ROI    &37.5 &57.8 & 40.9\\
         Bounding box & Bounding box       &\textbf{38.9} & \textbf{58.2} & \textbf{42.5} \\ 
        \bottomrule
    \end{tabular} 
    \label{table:features}
\end{center}
\end{table}

\noindent\textbf{Compared With Baselines.}~To show the effectiveness of the proposed DIR. We conduct experiments by integrating DIR with popular two stage detectors: Faster R-CNN, Mask R-CNN and Cascade R-CNN and compare results with several existing NMS methods. From Table ~\ref{table:compare_with_baselines} we can see that by applying DIR to these methods, the performance improved by 2.5, 2.2 and 0.8 respectively. The results also show that our method has a greater boost on the AP 75 than the AP 50, even in the strong baseline Cascade R-CNN, the AP 75 is boosted 1.6 in AP. We also integrate our DIR with several strong 
baselines: DCN, Double-Head and TSD. As shown in Table ~\ref{table:compare_with_strong_baselines}, our DIR can boost these methods by 1.8, 1.5 and 1.7 in AP respectively.

Both of our DIR and Cascade R-CNN work in a cascade manner, but the motivation is quite different. Cascade R-CNN proposes to refine bounding boxes by multiple cascade detection heads under increasing IoU threshold. While the purpose of the extra stage in DIR is to correct the feature misalignment and the instability of ground-truth in IoU regression. Faster R-CNN with DIR outperforms two stage Cascade R-CNN~\cite{cascadercnn} (38.9 vs 38.2 in AP) which further demonstrate the superiority of the predicted IoU in DIR over the classification scores.

The classification scores in Faster R-CNN have a strong ability to distinguish whether the bounding box's IOU is greater than 0.5, which is also the training target of the classifier. However, as shown in Figure ~\ref{fig:compare_p}, the classification score can't accurately measure the localization confidence of the bounding boxes which IoU is greater than 0.5. The Pearson correlation coefficient of IoU with ground-truth and the classification score is only 0.209. The proposed method can acquire accurate localization confidence of bounding boxes by predicting IoU with ground-truths. Figure ~\ref{fig:compare_p} shows that our method obtains 0.755 Pearson correlation coefficient, which significantly outperforms the Faster R-CNN baseline. These results further verify the effectiveness of the proposed DIR model.

%\begin{table*}[ht]
%\begin{center}
%    \caption{Influence of combination method of Purity and Integrity, Integrity and Purity indicate using Integrity or Purity as %the localization confidence. Geometric Average and arithmetic Arithmetic means using these two kind of average of Purity and %Integrity as localization confidence. Combined IoU indicates using the IoU obtained by Eq.\ref{eq:pure_inter_to_iou} as %localization confidence.}
%    \begin{tabular}{l c c c c c c}
%         \toprule
%         Method & AP & \( AP^{50} \) &\(AP^{75}\) & \(AP^{S}\) & \(AP^{M}\) & \(AP^{L} \) \\ \hline
%         Faster R-CNN                                    &36.4 & 58.0 & 39.3 & 21.4 & 40.3 &46.7  \\
%         Faster R-CNN + Integrity                        &37.7 & 59.3 & 41.6 & 22.2 & 41.9 &48.4  \\
%         Faster R-CNN + Purity                           &37.9 & 58.3 & 41.8 & 22.4 & 41.9 &49.1  \\
%         Faster R-CNN + Geometric Average                &38.7 & 58.4&42.2&22.5&42.7&50.4  \\   
%         Faster R-CNN + Arithmetic Average               &38.7 & 58.4&42.2&22.5&42.8&50.4  \\   
%         Faster R-CNN + Combined IoU                              &38.9 & 58.2 & 42.5 & 22.6&42.9&51.0   \\
%         \bottomrule
%    \end{tabular} 
%    \label{table:combine}
%\end{center}
%\end{table*}

%\setlength{\tabcolsep}{2pt}

\begin{table}[ht]
\begin{center}
    \caption{Influence of combination method of Purity and Integrity. Geometric Average and arithmetic Arithmetic means using these two kind of average of Purity and Integrity as localization confidence. Combined IoU indicates using the IoU obtained by Eq.\ref{eq:pure_inter_to_iou} as localization confidence.}
    \begin{tabular}{l c c c c c c}
         \toprule
         Method & AP & \( AP^{50} \) &\(AP^{75}\)  \\ \hline
         Faster R-CNN                                    &36.4 & 58.0 & 39.3  \\
         Faster R-CNN + Integrity                        &37.7 & \textbf{59.3} & 41.6  \\
         Faster R-CNN + Purity                           &37.9 & 58.3 & 41.8  \\
         Faster R-CNN + Geometric Average                &38.7 & 58.4 &42.2  \\   
         Faster R-CNN + Arithmetic Average               &38.7 & 58.4 &42.2  \\   
         Faster R-CNN + Combined IoU                     &\textbf{38.9} & 58.2 &\textbf{42.5} \\
         \bottomrule
    \end{tabular} 
    \label{table:combine}
\end{center}
\end{table}

\noindent\textbf{Influence of IoU Prediction with Hindsight.}~To validate the effectiveness of IoU Prediction with hindsight, we build a \textit{Foresight IoU Regressor} which predict the IoU of bounding boxes through the features of proposals. Other settings are the same as the Hindsight IoU regressor. As can be seen in Table \ref{table:ablation}, Faster R-CNN with Foresight IoU Regressor (2nd row) achieves 37.2\% in AP, 0.8\% higher than the baseline. Faster R-CNN with Hindsight IoU Regressor (4th row) achieves 38.3\% in AP, which is 1.9\% higher than the baseline and 1.1\% higher than the Foresight IoU Regressor. In terms of AP under high IoU standards, Hindsight IoU Regressor outperforms Foresight IoU Regressor 1.4\% in \(AP^{75}\), which demonstrates that IoU prediction with hindsight is more effective than with foresight. Figure ~\ref{fig:compare_p} also shows that the IoU predicted by hindsight manner have a higher Pearson correlation coefficient than the foresight manner (0.745 vs 0.633).

We also conduct experiments to explore the influence of features used for training and testing on the performance of IoU prediction. As shown in Table \ref{table:features}, whether for inference or training, using features of proposals to predict the IoU of the bounding boxes will seriously affect the performance, which shows that predicting IoU with hindsight is very effective and necessary. From this table, we can also see that using the features of bounding boxes for inference has a greater impact on performance. Even using features of proposals for training and using features of bounding boxes for inference can improve AP by 0.9\% than baseline.

\noindent\textbf{Influence of Purity and Integrity.}~We conduct experiments to analyze the influence of obtaining IoU by predicting the Purity and Integrity. As shown in Table \ref{table:ablation}, by replacing directly IoU prediction with predicting Purity and Integrity, both the Foresight IoU Regressor and Hindsight IoU Regressor have been steadily improved. The Foresight IoU Regressor has been improved from 37.2\% to 37.9\%. The Hindsight one has been improved from 38.3\% to 38.9\%. Figure ~\ref{fig:compare_p} also shows that decoupling IoU into Purity and Integrity improves the correlation coefficient between the ground-truth IoU and the predicted IoU from 0.743 to 0.755.

Since we use the sum of \(L_{purity}\), \(L_{integrity}\) and \(L_{IoU}\) to train the IoU predictor, to demonstrate that the improvement is not caused by the increment of learning rate, we triple the learning rate of a traditional IoU Regressor which obtain IoU by directly predict it for comparison. As shown in the 5th row of Table.\ref{table:ablation}, although the learning rate has been increased by three times, the performance has only improved slightly. 

We also conduct experiments to explore the effects of Purity and Integrity on the final localization confidence and try other methods to combine Purity and Integrity besides Eq.\ref{eq:pure_inter_to_iou}. As shown in Table \ref{table:combine}, only leveraging Purity or Integrity alone can also improve performance, but the best results are obtained when those two are used together. As for the combination method, Table \ref{table:combine} shows that combining Purity and Integrity by Eq.\ref{eq:pure_inter_to_iou} outperforms others method, which is also consistent with the definition of IoU and demonstrates that the IoU decomposition approach in our method is essential and effective.

\noindent\textbf{Inference Time.}~
Compared with the baseline model, adding additional IoU prediction modules will increase the time for inference. We test inference FPS of various settings with 8 Tesla V100 GPUs. As can be seen in Table \ref{table:ablation}, adding additional regressors (2nd row) will reduce the inference FPS by 1. Replacing the Foresight Regressor with Hindsight Regressor will reduce the fps by 0.4, but will increase the AP by 1.1. Replacing directly IoU prediction with predicting of Purity and Integrity will almost not affect inference time while steadily increasing AP.

\begin{table*}[t!]
\begin{center}
    \caption{Comparison with the state-of-the-art single-model detectors on MS COCO test-dev, * denotes using \(2\times\) training setting, The total number of training iterations will be doubled in the \(2 \times \) setting.}
    \begin{tabular}{c c   c  c c  c c  c}
    \toprule
    Method & Backbone  & \(AP\)& \( AP^{50} \) &\(AP^{75}\) & \(AP^{S}\) & \(AP^{M}\) & \(AP^{L} \) \\
    \midrule 
    Faster R-CNN~\cite{fpn} & ResNet-101-FPN &36.2 &59.1 &39.9&18.2&39.0&48.2 \\
    Faster R-CNN~\cite{grmi} &Inception-ResNet-v2-TDM & 36.8& 57.7& 39.2& 16.2& 39.8& 52.1 \\
    Mask R-CNN~\cite{maskrcnn} & ResNet-101-FPN &38.2 &60.3& 41.7& 20.1& 41.1& 50.2 \\
    Cascade R-CNN~\cite{cascadercnn} & ResNet-101-FPN& 42.8& 62.1& 46.3& 23.7& 45.5 &55.2\\
    Fitness NMS~\cite{fit} & DeNet-101~\cite{denet} & 41.8 &60.9& 44.9& 21.5 &45.0& 57.5\\
    Deformable R-FCN~\cite{deformable} & Aligned-Inception-ResNet& 37.5& 58.0& 40.8& 19.4& 40.1& 52.5 \\
    IoU-Net*~\cite{iounet} & ResNet-101-FPN &40.6& 59.0& - &- &- &-\\
    Rank-NMS~\cite{ltr} & ResNet-101-FPN &41.0 &60.8 &44.5& 23.2& 44.5& 52.5 \\ 
    HTC~\cite{chen2019hybrid} & ResNeXt-101-FPN-DCN &50.7 &70.5  &55.2 &32.0 &53.8 &64.0 \\
    
    \midrule
    Faster R-CNN + Ours& ResNet-101-FPN &41.1 & 60.5 &44.6 &23.4 & 44.4 &52.0 \\
    Mask R-CNN + Ours & ResNet-101-FPN &41.6 & 60.9&45.4 &23.7 &45.0 & 52.8  \\
    Faster R-CNN* + Ours& ResNet-101-FPN &41.9 &61.0 &45.5&23.3&45.4&53.4\\
    Mask R-CNN* + Ours& ResNet-101-FPN &42.6 & 61.7 & 46.3 & 23.7 & 46.0 & 54.8   \\
    HTC + Ours & ResNeXt-101-FPN-DCN &51.3 &70.5 &55.5 &32.5 &54.7 &65.2 \\
    \bottomrule
    \end{tabular} 
    \label{table:sota}
\end{center}
\end{table*}
\subsection{Comparison with state-of-the-art Detectors}~We compare the proposed method with state-of-the-art object detection models on MS COCO test-dev. By integrating with our method, the detection performance of Faster R-CNN, Mask R-CNN and HTC has been significantly improved and achieve state-of-the-art. As shown in Table \ref{table:sota}, compared with IoU-Net which also introduces an IoU predictor to get the localization confidence, our method outperforms it by a large margin with more concise design. Our method also outperforms the state-of-art NMS method Rank-NMS~\cite{ltr} which acquire the localization confidence of bounding boxes by learning to rank proposals. These results validate the superiority of the proposed Hindsight IoU Regressor.

\begin{figure*}[t]
  \begin{center}
  \includegraphics[height=0.40\linewidth]{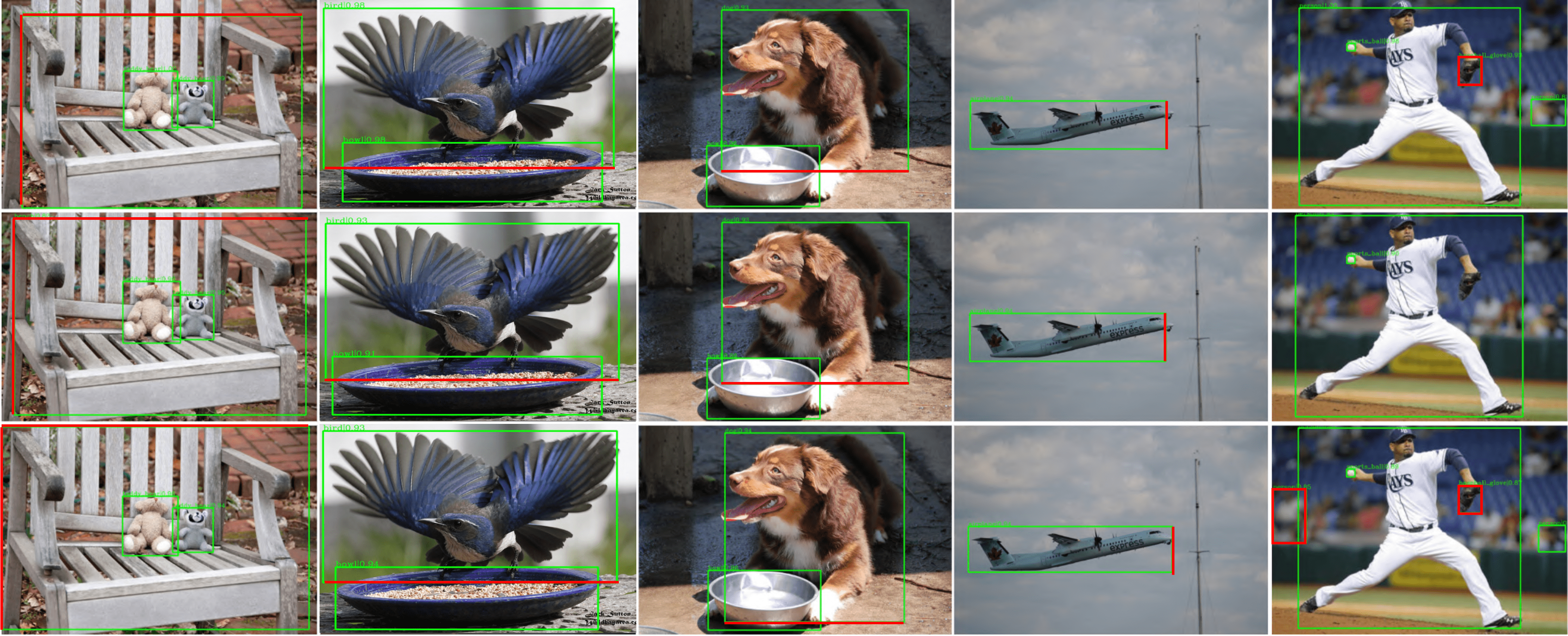}
  \caption{The visualizations of detect results of different methods. The first row are the results of Faster R-CNN, the second row are the results of Foresight IoU Regressor, the last row are the results of the proposed DIR-Net . The difference between the detection boxes is highlighted in red color.}
  \label{fig:compare}   
  \end{center} 
\end{figure*} 

\subsection{Qualitative results}~Qualitative results for comparison between the proposed DIR-Net with Faster R-CNN and the Foresight IoU Regressor are provided in Figure \ref{fig:compare}. As shown, our method can obtain more accurate object boundaries compared with Faster R-CNN and the Foresight Regressor. Moreover, our method is able to recall more objects than the foresight one. This demonstrates that the proposed method can acquire more reliable localization confidence of bounding boxes. 

\section{Conclusion}
In this paper, we propose a novel DIR model to accurately evaluate the localization confidence of detected bounding boxes. 
By analyzing the definition of IoU, we find that IoU mathematically entangling Purity and Integrity that rely on different information. 
Compared with directly predicting IoU, the proposed DIR uses two sub-network branches to separately model Purity and Integrity, and then combine them to get IoU. 
Such decoupled manner can divide the complex mapping between the bounding box and its IoU into two easier mappings. In addition, we analyze the instability and feature misalignment caused by foresight IoU regression. A simple but effective feature realignment approach is introduced to make the IoU regression model work in a hindsight manner, which makes the target mapping that the model needs to learn more stable. Experiments show that the proposed DIR can be integrated with popular two-stage detectors and significantly improve their detection performance.

\section*{ACKNOWLEDGMENTS}
This work was supported by Alibaba Group through Alibaba Research Intern Program.
\bibliographystyle{ACM-Reference-Format}

\bibliography{sample-base}

\end{document}